%% file: acl_latex.tex
\pdfoutput=1
\documentclass[11pt]{article}

\usepackage[]{acl}

\usepackage{times}
\usepackage{latexsym}

\usepackage[T1]{fontenc}
\input{math_commands.tex}

\usepackage[utf8]{inputenc}

\usepackage{microtype}

\usepackage{inconsolata}

\usepackage{graphicx}
\usepackage{booktabs}
\usepackage{multirow}
\usepackage{xcolor} 
\usepackage{listings}
\usepackage{hyperref}
\usepackage{multicol} % for multiple columns
\usepackage{lipsum} % for dummy text
\usepackage{amsmath}
\usepackage{amssymb}
\usepackage{amsfonts} 
\usepackage{algorithm}

\usepackage{algpseudocode}
\usepackage{array}
\usepackage{booktabs}
\usepackage{arydshln}
\title{CogLM: Tracking Cognitive Development of Large Language Models}

\author {
    \textbf{Xinglin Wang}\textsuperscript{\rm 1}\footnotemark[1], \hspace{0cm}
    \textbf{Peiwen Yuan}\textsuperscript{\rm 1}\footnotemark[1], \hspace{0cm}
    \textbf{Shaoxiong Feng}\textsuperscript{\rm 2}, \hspace{0cm} 
    \textbf{Yiwei Li}\textsuperscript{\rm 1}, \hspace{0cm} 
    \textbf{Boyuan Pan}\textsuperscript{\rm 2}, \hspace{0cm} \\
    \textbf{Heda Wang}\textsuperscript{\rm 2}\textbf{,} \hspace{0cm} 
    \textbf{Yao Hu}\textsuperscript{\rm 2}\textbf{,} \hspace{0cm} 
    \textbf{Kan Li}\textsuperscript{\rm 1}\footnotemark[2] \\
    \textsuperscript{\rm 1} School of Computer Science, Beijing Institute of Technology \\
    \textsuperscript{\rm 2} Xiaohongshu Inc \\
    \texttt{\{wangxinglin,peiwenyuan,liyiwei,likan\}@bit.edu.cn} \\
    \texttt{\{shaoxiongfeng2023,whd.thu\}@gmail.com} \  \texttt{\{panboyuan,xiahou\}@xiaohongshu.com}
}

\begin{document}
\maketitle

\renewcommand{\thefootnote}{\fnsymbol{footnote}} 
\footnotetext[1]{Equal contribution.} 
\footnotetext[2]{Corresponding author.}
\renewcommand{\thefootnote}{\arabic{footnote}}

\input{abstract}
\input{introduction}

\input{method}
\input{experiment}

\input{related_work}

\input{conclusion}
\input{limitation}
\input{ethic}
\input{acknowledgement}

% \section*{Acknowledgements}
% \clearpage

% Entries for the entire Anthology, followed by custom entries
\bibliography{anthology,custom}
% \bibliography{custom}
\clearpage

\appendix
\input{appendix}

\end{document}

%% file: math_commands.tex
\usepackage{amsmath,amsfonts,bm}

\def\eqref#1{equation~\ref{#1}}
\def\1{\bm{1}}

\DeclareMathAlphabet{\mathsfit}{\encodingdefault}{\sfdefault}{m}{sl}
\SetMathAlphabet{\mathsfit}{bold}{\encodingdefault}{\sfdefault}{bx}{n}

\def\sS{{\mathbb{S}}}

%% file: abstract.tex
\begin{abstract}
Piaget's Theory of Cognitive Development (PTC) posits that the development of cognitive levels forms the foundation for human learning across various abilities.
As Large Language Models (LLMs) have recently shown remarkable abilities across a wide variety of tasks, we are curious about the cognitive levels of current LLMs: to what extent they have developed and how this development has been achieved. 
% However, the mechanism regarding why and how such performance has been achieved remains unknown, and it is unclear whether LLMs can achieve human-like cognitive abilities or whether these models are still fundamentally limited. 
% To bridge this gap, we introduce Piaget's Theory of Cognitive Development (PTC) as a tool to reveal the development of cognitive abilities of LLMs. 
To this end, we construct a benchmark \textsc{CogLM} (\textbf{Cog}nitive Ability Evaluation for \textbf{L}anguage \textbf{M}odel) based on PTC to assess the cognitive levels of LLMs. \textsc{CogLM} comprises 1,220 questions spanning 10 cognitive abilities crafted by more than 20 human experts, providing a comprehensive testbed for the cognitive levels of LLMs. 
Through extensive experiments across multiple mainstream LLMs with \textsc{CogLM}, we find that: (1) In our testing framework, advanced LLMs (such as GPT-4) have demonstrated human-like cognitive abilities, comparable to those of a 20-year-old human. (2) The parameter size and optimization objective are two key factors affecting the cognitive levels of LLMs. (3) The performance on downstream tasks is positively correlated with the level of cognitive abilities. 
These findings fill the gap in research on the cognitive abilities of LLMs, tracing the development of LLMs from a cognitive perspective and guiding the future direction of their evolution.\footnote{Our code and data have been released on \url{https://github.com/WangXinglin/CogLM}.}
% These findings provide guidance for the future development of advanced abilities of LLMs from the perspective of ability evolution, and shed light on the mystery behind the emergence of advanced abilities of LLMs.
\end{abstract}

%% file: introduction.tex
\section{Introduction}
Large Language Models (LLMs) have recently achieved impressive performance on a wide variety of Natural Language Processing (NLP) tasks, including text comprehension \citep{kenton2019bert}, reasoning \citep{talmor2020leap, webb2023emergent}, code generation \citep{chen2021evaluating}, and mathematical problems \citep{fu2023chain}. 
% However, there is little 
% theoretical evidence regarding why and how this performance has been achieved. While recent research focuses on evaluating or enhancing the upper bound of LLMs' capabilities on certain type of tasks \citep{plan, abstract_reason}, few 
However, few studies have explored the reasons behind the evolutionary relationship among various abilities, which makes it difficult to understand the development of LLMs' capabilities as a whole and may pose potential risks to their further development.

% 我们期望去衡量meta能力，以及这些能力给llm带来哪些愿景  （ 对meta能力的度量有助于理解当前语言模型在复杂任务上的外在表现，指引LLM认知能力的未来提升方向）。关键词meta-ability
% 为了去研究meta能力，引入piaget，给出piaget的定义、四阶段说明与meta-ablility联结。
To this end, we introduce Piaget's Theory of Cognitive Development (PTC) \citep{piaget, 1977cognitive, childrenevaluation}, which posits that the development of cognitive levels forms the foundation for human learning across various abilities. 
Inspired by this, we think that studying the cognitive development of LLMs can assist us in better understanding the current performance of LLMs on downstream tasks and illuminate the path for future enhancements of their capabilities.
% Inspired by the recent success of cognitive psychology research on LLMs \citep{usingcognitive,ToM}, we introduce Piaget's Theory of Cognitive Development (PTC) \citep{piaget, 1977cognitive, childrenevaluation} as a tool to evaluate the cognitive abilities of LLMs. 
As the most authoritative theory in the development of psychology, PTC suggests that human children move through four different stages of learning, including the sensorimotor stage (0-2 years old), the preoperational stage (2-7 years old), the concrete operational stage (7-12 years old), and the formal operational stage (above 12 years old). Children in different cognitive stages exhibit significantly distinct patterns of thinking and cognitive abilities, which affects their learning of other skills. 
% 过渡一下
Examining LLMs from the perspective of PTC, some natural and crucial questions are: At what stage has the cognitive ability of LLMs developed compared to humans at present? What are the key factors that affect the cognitive abilities of LLMs? Is the emergence of advanced abilities and performance bottlenecks in current LLMs related to their cognitive levels?
% does the development of LLMs' abilities also follow similar patterns as human cognitive development?
%Just like a 5-year-old child cannot understand advanced mathematics.
% PTCD focuses not only on understanding how children acquire knowledge, but also on understanding the nature of intelligence. 

To explore the above questions, we construct a benchmark based on the scenario experiments used in PTC for evaluating the cognitive abilities of LLMs, denoted as \textsc{CogLM}. A large-scale human trial was conducted involving 207 participants aged between 6 and 20 years to ensure the alignment between the \textsc{CogLM} and PTC.
 We then perform extensive experiments on \textsc{CogLM} over several series of language models, including OPT \citep{opt}, Llama-2 \citep{llama2}, GPT-3.5-Turbo and GPT-4 \citep{gpt4}. Our results indicate that: (1) Under our testing framework, Advanced LLMs, such as GPT-4, have developed human-like cognitive abilities, matching those of a 20-year-old individual. (2) Two primary factors influencing these cognitive capacities in LLMs are the size of their parameters and their optimization objectives. (3) There is a positive correlation between the cognitive level of LLMs and their performance in downstream tasks.
% (1) As the size of parameters in LM increases, its cognitive ability shows an upward trend, similar to the aging process in humans. 
% （3）LLMs的认知能力与其高级能力的联系 （复杂、困难任务上的表现）
%Is the capability development of LLMs consistent with the process of human cognitive development? Is there a developmental sequence in the abilities of LLMs (similar to human)?

We believe that our findings can present a clear understanding of the current cognitive level of LLMs and provide insights into the emergence of advanced abilities in LLMs, shedding light on the future development of them. Our contributions can be summarised as follows:
\begin{itemize}
\item[$\bullet$] We innovatively introduce Piaget's Theory of Cognitive Development (PTC) to analyze the development of cognitive abilities of LLMs.
\item[$\bullet$] We construct a high-quality benchmark (\textsc{CogLM}) for evaluating the cognitive ability level of LLMs.
\item[$\bullet$] Comprehensive experiments across multiple LLM series on \textsc{CogLM} demonstrate the cognitive level of current LLMs, key factors responsible for their varying levels, and relationships between cognitive levels and performance on downstream tasks.
\end{itemize}

%Despite these early limitations, the question of whether the emergence of LLMs' abilities coincide with the process of human cognitive development remains open, as measuring the level of intelligence is a challenging task.

%% file: method.tex
\section{\textsc{CogLM} Benchmark Development}
% 为了全面准确衡量语言模型的认知能力，我们进行了以下努力：（1）我们revisit了皮亚杰认知理论所包含的12项认知能力，结合语言模型的特点，我们从中选择并re定义了10项能力构建\textsc{CogLM}（第一节）。（2）我们创建了标准化数据构建指南，从而保障\textsc{CogLM}的质量。（第二节）（3）我们在\textsc{CogLM}上进行了广泛的人类测试，并结合皮亚杰认知理论，确保\textsc{CogLM}能够准确反映人类的认知水平。（第三节）（4）我们构建了Calibrated Mapping Function，获得可信的\textsc{CogLM}答题情况到认知年龄的映射关系。（第四节）

% Constancy, Early Representation, Semiotic Function, Self-Center,  Reversibility,  Conservation, Induction, Hypothetico-Deductive, Propositional Operation, Plan

To comprehensively and accurately assess the cognitive abilities of LLMs, we undertake the following efforts:
% (1) \textbf{Revisiting Piaget's Cognitive Theory:} We first revisit 12 cognitive abilities proposed by PTC, 10 of which are selected and redefined to construct \textsc{CogLM} according to the characteristics of LMs (section \ref{sec:3.1}).
% (2) \textbf{Developing Standardized Data Construction Guidelines:} We create standardized data construction guidelines to ensure the quality of \textsc{CogLM} (section \ref{sec:3.2}).
% (3) \textbf{Conducting Human Testing:} We conducted extensive human testing to ensure the alignment between \textsc{CogLM} and PTC (section \ref{sec:3.3}).
% (4)\textbf{ Constructing Calibrated Mapping Function:} We built a Calibrated Mapping Function to establish a reliable mapping between testing results on \textsc{CogLM} and cognitive age (section \ref{sec:3.4}).
(1) We revisit 12 cognitive abilities proposed by PTC, 10 of which are selected and redefined to construct \textsc{CogLM} according to the characteristics of LLMs (section \ref{sec:3.1}).
(2) We create standardized data construction guidelines to ensure the quality of \textsc{CogLM} (section \ref{sec:3.2}).
(3) We conduct extensive human testing to ensure the alignment between \textsc{CogLM} and PTC (section \ref{sec:3.3}).
(4) We build a Calibrated Mapping Function to establish a reliable mapping between testing results on \textsc{CogLM} and cognitive age (section \ref{sec:3.4}).

\subsection{Definition of Cognitive Abilities}
\label{sec:3.1}
According to PTC, the development of human cognition is divided into four stages, which include 12 cognitive abilities. Considering that the interaction interface of most LMs is limited to text-based format, we exclude \textit{reflexes} and \textit{sensorimotor} aspects of multimodal interaction and build \textsc{CogLM} based on the remaining 10 cognitive abilities.

We strictly define the concept of each cognitive ability based on PTC and provide representative examples for explanation as shown in Table \ref{table:def}.

\begin{table*}[t]
\small
\centering
\begin{tabular}{p{13.5cm}}
\toprule
\textbf{First Stage: \textit{Constancy} (\textit{const})} \\
\textbf{Definition:} Objects exist both within and outside the field of vision and maintain a certain level of stability. \\
\textbf{Example: } Q: Assuming there is a small ball on the table. Is the ball still on the table when covered with a cloth?   Ans: Yes \\
\midrule
\textbf{First Stage: \textit{Early Representation} (\textit{early}) } \\
\textbf{Definition:} Objects are endowed with corresponding meanings, thus gradually forming a universe of permanent objects. \\
\textbf{Example: } Q: How would you describe the color of snow?   Ans: White \\
\midrule
\textbf{Second Stage: \textit{Semiotic Function} (\textit{semio})} \\
\textbf{Definition:} Use symbols to represent things and concepts. \\
\textbf{Example: } Q: Which item best represents love and romance?   Ans: Rose \\
\midrule
\textbf{Second Stage: \textit{Empathy} (\textit{empat})} \\
\textbf{Definition:} Start considering others' perspectives and feelings when addressing issues. \\
\textbf{Example: } Q: You are fond of video games, but your cousin enjoys outdoor sports. What birthday gift would you give to him?   Ans: Camping tent \\
\midrule
\textbf{Third Stage: \textit{Reversibility} (\textit{rever})} \\
\textbf{Definition:} Understand the reversibility of physical operations and is capable of reverse thinking. \\
\textbf{Example: } Q: If a plane lands at 10 AM and flies for 6 hours, what time will it take off?   Ans: 4 AM \\
\midrule
\textbf{Third Stage: \textit{Conservation} (\textit{conse})} \\
\textbf{Definition:} The external alteration of forms doesn't affect certain fundamental attributes. \\
\textbf{Example: } Q: If a stone is cut into two, what will be their total mass?   Ans: No change \\
\midrule
\textbf{Third Stage: \textit{Induction} (\textit{induc})} \\
\textbf{Definition:} Infer universal rules based on observed results. \\
\textbf{Example: } Q: Given an arithmetic sequence: 2, 5, 8, 11, 14, which of the following is the general term formula for this sequence?   Ans: 3n + 2 \\
\midrule
\textbf{Forth Stage: \textit{Hypothetico-Deductive} (\textit{deduc})} \\
\textbf{Definition:} Deduce practical problems based on specific assumptions or rules. \\
\textbf{Example: } Q: Alex is excited, Paul is sad, Mike is crying, Anna is angry. The sad one is a dog, the angry one is swan, the crying one is cat, the excited one is tiger. Swan likes cats, cat likes tiger, tiger likes dog, dog likes swan. What does Anna like?   Ans: Cat \\
\midrule
\textbf{Forth Stage: \textit{Propositional Operation} (\textit{propo})} \\
\textbf{Definition:} Understand propositions and determine the logical relationships between propositions. \\
\textbf{Example: } Q: Sentence1: In fact, the Lions of Delos were made from Naxos marble. Sentence2: There are five Lions of Delos, and also two Tigers of Delos. What is the propositional relationship between sentence1 and sentence2 ?   Ans: Neutral \\
\midrule
\textbf{Forth Stage: \textit{Plan} (\textit{plan})} \\
\textbf{Definition:} Develop sensible solutions based on specific problem. \\
\textbf{Example: } Q: Please plan an action execution sequence according to the rules. The following rules must be followed: going fishing before going hiking, doing yoga before going hiking, taking photos before doing yoga. Based on the above rules, please choose an action execution sequence that meets the rules from the following options to finalize: going hiking.   Ans: taking photos, doing yoga, going fishing, going hiking \\
\bottomrule
\end{tabular}
\caption{Definitions and examples of cognitive abilities included in \textsc{CogLM}.}
\label{table:def}
\vspace{0cm}
\end{table*}

% 依据PTC，人类的认知能力主要包括Sensorimotor Stage：，Preoperational Stage，Concrete Operations Stage，Formal Operations Stage
% 考虑到LMs的交互接口局限于文本形式，我们将需要多模态交互的reflexes和sensorimotor排除在外，并用余下十种认知能力构建\textsc{CogLM}。
% 从同化和顺应演化而来，cognitive ability（定义加例子）

% 为了构建出

\subsection{Standardized Annotation Guidelines}
\label{sec:3.2}
% 为了确保所要构建的数据集能够准确反映语言模型的认知能力，我们在正式标注前制定了standardized data construction guidelines，并在标注阶段严格执行该guidelines:
% 数据类型：尽管modern大语言模型普遍有着良好的生成能力，但是早期的（early-aged） LMs (e.g., GPT2)的生成能力较为有限，因此我们采用了选择题作为考察形式。这能够避免生成能力的差异性影响对认知能力的准确评估。该设定也与人类在婴幼儿时期语言、写作能力欠缺，因此适合采用选择的方式来考察认知能力的惯例一致。
% 题目数量：早期阶段的能力比较简单、体现形式比较集中，而晚期阶段的能力比较综合、体现形式比较多样。基于此，我们采用了随阶段延伸，题目数量由少到多的设定，如表所示。
% 标注者培训：我们选择了具有心理学背景或人工智能教育背景的成年人作为数据标注者。我们整理了有关PTC的系统资料，并要求标注者认真学习。之后我们通过试卷（试卷内容见Appendix）考察的方式确保标注者充分理解了PTC相关背景知识。最后，我们将Table 1中的资料作为标注指南，并为每个认知能力提供不少于三道例题。每名标注者被要求分别为两个特定认知能力标注不少于30道题目及选项。
% 标注质量控制：标注结束后，我们通过标注者交叉检查的方式筛选掉存在质量隐患的例题。质量隐患包括：题目内容不能有效评估相应的认知能力、歧义的存在使得题目答案不合理、存在偏见暴力等元素。
To ensure that \textsc{CogLM} can accurately reflects the cognitive abilities of LLMs, we have established standardized annotation guidelines and strictly adhere to them during the annotation phase:

\textbf{Data Format} Although modern LLMs generally possess strong generation capabilities, early-aged LLMs (e.g., GPT-2) have limited generation abilities (similar to Human infants). Therefore, we have opted for multiple-choice questions as the assessment format. This approach avoids the influence of variations in generation capabilities on the accurate evaluation of cognitive abilities.

\textbf{Number of Samples} Abilities in the early stages are relatively simple and have a more concentrated form of expression, while abilities in the later stages are more comprehensive and have a more diverse form of expression. Based on this, we have set the number of samples to increase with each stage, as shown in the Table~\ref{tab: statistics}.

\textbf{Qualified Annotator} We select adults with backgrounds in psychology or artificial intelligence as data annotators. Annotators are provided with detailed materials on PTC and required to study them carefully. We then assess annotators' understanding of PTC through exams (see Appendix Table \ref{table:appendix-exam} for the examination paper). Finally, we provide annotators with at least 3 example samples for each cognitive ability. Each annotator is required to annotate no fewer than 30 questions and options for two specific cognitive abilities.

\textbf{Annotation Quality Control} After annotation, we conduct cross-checks among annotators to identify samples with quality issues. Quality issues include questions that cannot effectively assess the corresponding cognitive abilities, questions with ambiguities, and elements of bias or violence.

\subsection{Consistency with Theory}
\label{sec:3.3}
% 在数据集构建完成后，我们考虑通过人类测试进一步确定\textsc{CogLM}是否与PTC一致，能否有效反应认知能力。我们通过从\textsc{CogLM}的每个subset中随机抽取10条数据组成问卷，并发放给6-18岁的测试者完成测试。测试情况的详细统计见Appendix。在235份答卷中，有187份被认定为有效（依据答题时间是否合理）。我们统计了受试者的年龄和答卷得分的斯皮尔曼与皮尔逊相关系数，表结果显示二者呈现了显著的相关性。该统计结果从侧面验证了我们所制定的Standardized Annotation Guidelines能够保证\textsc{CogLM}测试认知能力的有效性。
After the dataset construction is completed, we consider conducting human tests to further ascertain whether \textsc{CogLM} is consistent with PTC and whether it can effectively reflect cognitive abilities. We randomly select 10 samples from each subset of \textsc{CogLM} to create questionnaires, which are then distributed to testers aged between 6 and 20. Out of the 207 completed questionnaires, 141 are deemed valid (based on the reasonableness of test duration\footnote{We deem the papers completed by the questionnaires in less than 10 minutes as invalid. Humans received the same multiple-choice questions to answer as LLMs.}). We calculate the Spearman and Pearson correlation coefficients between the age of the participants and their questionnaire scores. It turns out that spearman correlation is 0.7169 and pearson correlation is 0.7362 (all the p-values $< 1e-10$), indicating a strong correlation between them.
% The results shown in Table \ref{tab:test} indicates a strong correlation between them. 
This statistical result validates the effectiveness of the Standardized Annotation Guidelines we have developed in ensuring the efficacy of \textsc{CogLM} for assessing cognitive abilities.

% Please add the following required packages to your document preamble:
% \usepackage{multirow}
\begin{table*}[t]
\setlength{\tabcolsep}{2pt} % 设置列之间的空白长度
\small
\centering
\renewcommand\arraystretch{1.025}
\begin{tabular}{cccp{0.001pt}ccp{0.001pt}cccp{0.001pt}cccc}
\toprule
\multirow{2}{*}{\textsc{CogLM}} & \multicolumn{2}{c}{stage 1} && \multicolumn{2}{c}{stage 2} && \multicolumn{3}{c}{stage3} && \multicolumn{3}{c}{stage4} & \multirow{2}{*}{Overall} \\ \cline{2-3}\cline{5-6}\cline{8-10}\cline{12-14}
 & const & early && semio & empat && rever & conse & induc && deduc & propo & plan &  \\ \hline
Sample Number & 50 & 100 && 100 & 100 && 100 & 110 & 100 && 250 & 100 & 210 & 1220 \\
Question Tokens & 18.5 & 11.36 && 11.55 & 25.27 && 30.0 & 26.3 & 42.0 && 51.8 & 30.0 & 77.9 & 39.5 \\
Candidates Number & 2.00 & 4.00 && 3.96 & 2.96 && 4.00 & 2.98 & 4.00 && 4.00 & 3.00 & 4.00 & 3.66 \\
Candidates Tokens & 1.00 & 1.19 && 1.48 & 4.23 && 3.87 & 4.28 & 7.58 && 1.00 & 1.00 & 20.30 & 5.71 \\
\bottomrule
\end{tabular}
\caption{Data statistics on all ability subsets of \textsc{CogLM}.}
\label{tab: statistics}
\end{table*}

\begin{table}[t]
\centering
\small
\renewcommand{\arraystretch}{1.2}
\setlength{\tabcolsep}{2pt}
\begin{tabular}{lcc}
\toprule
Type & Series & Size  \\ \midrule
\multirow{2}{*}{Text completion}&OPT & 125M, 1.3B, 2.7B, 6.7B  \\
\multirow{2}{*}{}&Llama-2 & 7B,13B,70B  \\ \midrule
\multirow{3}{*}{Chat completion}&Llama-2-chat & 7B,13B,70B \\
\multirow{2}{*}{}&GPT-3.5-Turbo & N/A   \\
\multirow{2}{*}{}&GPT-4 & N/A \\
\bottomrule
\end{tabular}
\caption{The statistics of considered language models.}
\label{tb:model}
\end{table}

% \begin{table}[t]
% \caption{ Correlation coefficients between the age of the participants and their questionnaire scores.}
% \centering
% \renewcommand\arraystretch{1.3}
% \begin{tabular}{cccc}
% \toprule
% Spearman correlation / p-value &  Pearson correlation / p-value\\ \midrule
% 0.7169  /  6e-10                 &     0.7362   /   9e-11      \\ 
% \bottomrule
% \end{tabular}
% \label{tab:test}
% \end{table}

% Human Eval：问卷，filter，（是否验证了数据集构建与理论的匹配性）

\subsection{Calibrated Cognitive Age Mapping Function}
\label{sec:3.4}
% 在明确了解答准确性与认知年龄的相关性后，我们希望进一步根据前者预估后者。为此，我们首先对准确性的计算方法进行了修正。\textsc{CogLM}中的题目的候选项分布在区间[2,5]。题目之间候选项数量的差异性会影响在受试者不确定答案时，通过猜测回答正确的可能性。因此我们通过以下公式来计算在某个subset S上的Calibrated accuracy：
After confirming the positive correlation between answer accuracy and cognitive age, we aim to further construct the mapping function between them. We first make adjustments to the method of calculating accuracy. The number of candidate options for questions in \textsc{CogLM} falls within the range $[2,5]$. Such a variability  can impact the likelihood of providing a correct answer through guessing when participants are uncertain. Therefore, we calculate the calibrated accuracy on certain subset $\sS$ as follows:

{\footnotesize
\begin{equation}
Acc = \frac{1}{|\sS|} \times \sum_{i=1}^{|\sS|} \frac{\1_\mathrm{predict_i = answer_i} - 1/|\mathrm{candidates_i}|}{1-1/|\mathrm{candidates_i}|}
\label{eq:cal}
\end{equation}
}

% 我们进一步以section \ref{sec:3.3}中的问卷结果为训练集 S_Q，优化回归函数f：
A negative calibrated accuracy (worse than random selecting) indicates a clear deficiency in the corresponding cognitive ability.
We further use 80\% of the questionnaire results in Section \ref{sec:3.3} as the training set $\sS_Q$ to optimize the regression function $f(\cdot)$ as follows:

{\footnotesize
\begin{equation}
\begin{gathered}
\mathcal{L}_{regression}= \frac{1}{|\sS_{Q}|} \times \sum_{i=1}^{|\sS_{Q}|}(f(Acc_i)-age_i)^2 \\
f(Acc) = \sum_{i=1}^{4} w_i \times Acc_{\mathrm{stage}i}+b
\end{gathered}
\label{equation: L_inter}
\end{equation}
}

The Spearman correlation between the age predicted by $f(\cdot)$ and the real age on the remaining 20\% samples is 0.9354, which signifies that $f(\cdot)$ can precisely approximate the mapping from results on \textsc{CogLM} to cognitive age. We observe that $w_1:w_2:w_3:w_4 = 1:2.6:1.4:2.5$, indicating that cognitive abilities in the second and fourth stages are better at reflecting cognitive age under the evaluation of \textsc{CogLM}. 

% etric构建：由人类数据，进行线性拟合，得到四个阶段得分到岁数的映射关系，作为metric

%% file: experiment.tex
\section{Experiments}
\subsection{Experimental Setup}

\begin{table*}[t]
\renewcommand{\arraystretch}{1.25}
\centering
\small
\setlength{\tabcolsep}{2.25pt} % 设置列之间的空白长度
\begin{tabular}{lccp{0.001pt}ccp{0.001pt}cccp{0.001pt}ccccc}
\toprule
\multicolumn{1}{c}{\multirow{2}{*}{Model}} & \multicolumn{2}{c}{stage1} && \multicolumn{2}{c}{stage2} && \multicolumn{3}{c}{stage3} && \multicolumn{3}{c}{stage4} & \multirow{2}{*}{Acc} & \multirow{2}{*}{Age} \\ \cline{2-3}\cline{5-6}\cline{8-10}\cline{12-14}
\multicolumn{1}{c}{} & const & early && semio & empat && conse & induc & rever && deduc & propo & plan &  \\ \hline
OPT 6.7B & -4.0 & 64.2 && 41.1 & -3.0 && 13.5 & 10.6 & 20.1 && -0.8 & -0.5 & 14.2 &  15.5 & 6.5 \\ 
Llama-2-chat-70B & 52.1 & 96.2 && 78.5 & 66.2 && 68.4 & 65.3 & 44.0 && 15.2 & 40.0 & 20.6 & 54.6 & 14.1 \\ 
GPT-3.5-Turbo & 92.0 & \underline{97.3} && 90.6 & \underline{85.5} && 65.9 & 64.0 & 61.3 && 27.5 & 49.0 & 6.7 & 64.0 & 16.1 \\ 
GPT-4 & \underline{96.0} & \underline{97.3} && \underline{96.0} & \textbf{90.3} && \underline{90.4} & \underline{78.7} & \underline{78.7} && \underline{99.4} & \underline{61.0} & \underline{59.4} & \underline{84.7} & \underline{20.0} \\ 
Human & \textbf{100.0} & \textbf{98.0} && \textbf{96.1} & 84.2 && \textbf{98.2} & \textbf{91.6} & \textbf{92.0} && \textbf{100.0} & \textbf{82.0} & \textbf{95.6} & \textbf{93.7} & \textbf{21.5} \\
\bottomrule
\end{tabular}
\caption{Calibrated accuracy (\%) of largest model in evaluating series. Acc and Age refer to calibrated accuracy and the age of equivalent human performance. The value of Age is calculated according to Equation~\ref{equation: L_inter}. Bold indicates the best performance.}
\label{tb:main}
\end{table*}

\begin{figure*}[t]
\begin{center}
\includegraphics[width=0.80\textwidth]{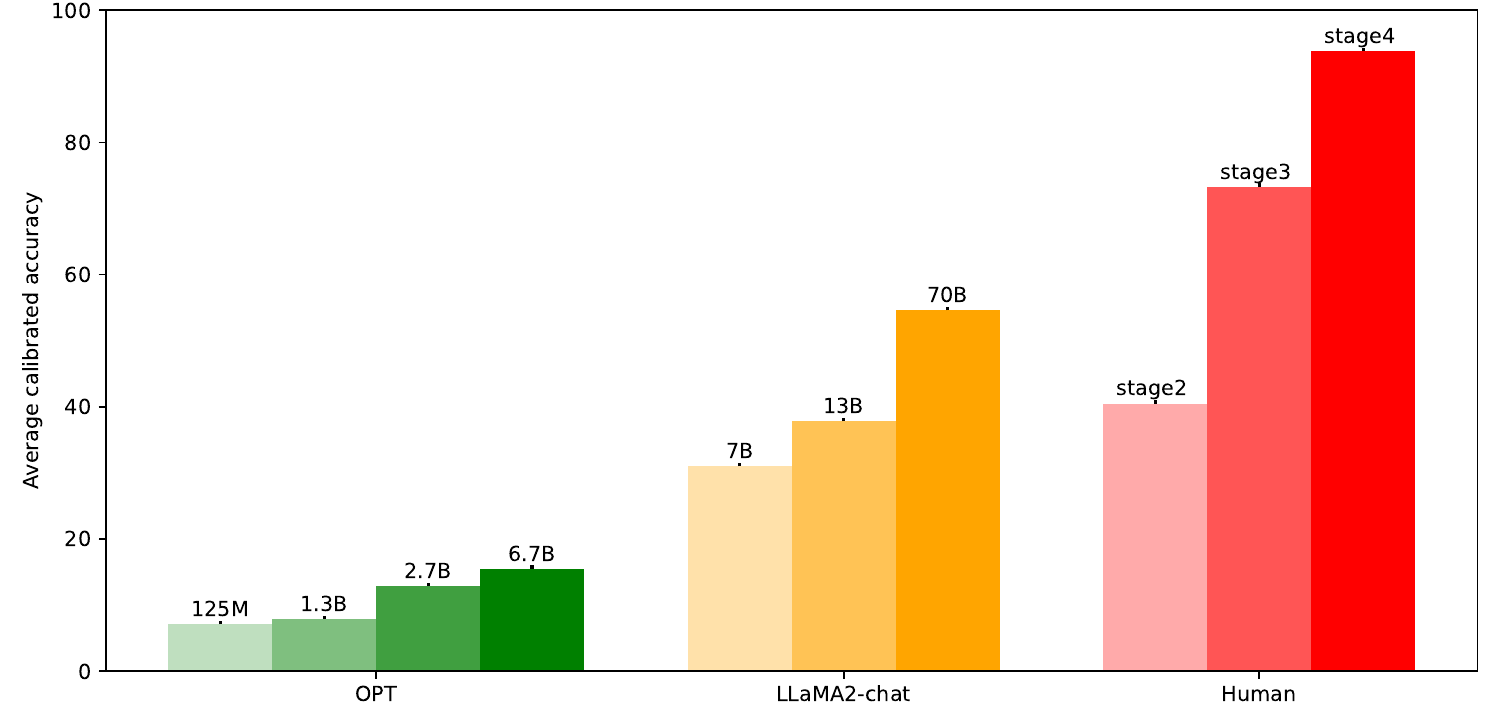}
\end{center}
\caption{Average calibrated accuracy (\%) of models with different parameter size and humans in different cognitive stage.}
\label{fig:parameter}
\end{figure*}

\begin{figure}[t]
\begin{center}
\includegraphics[width=0.44\textwidth]{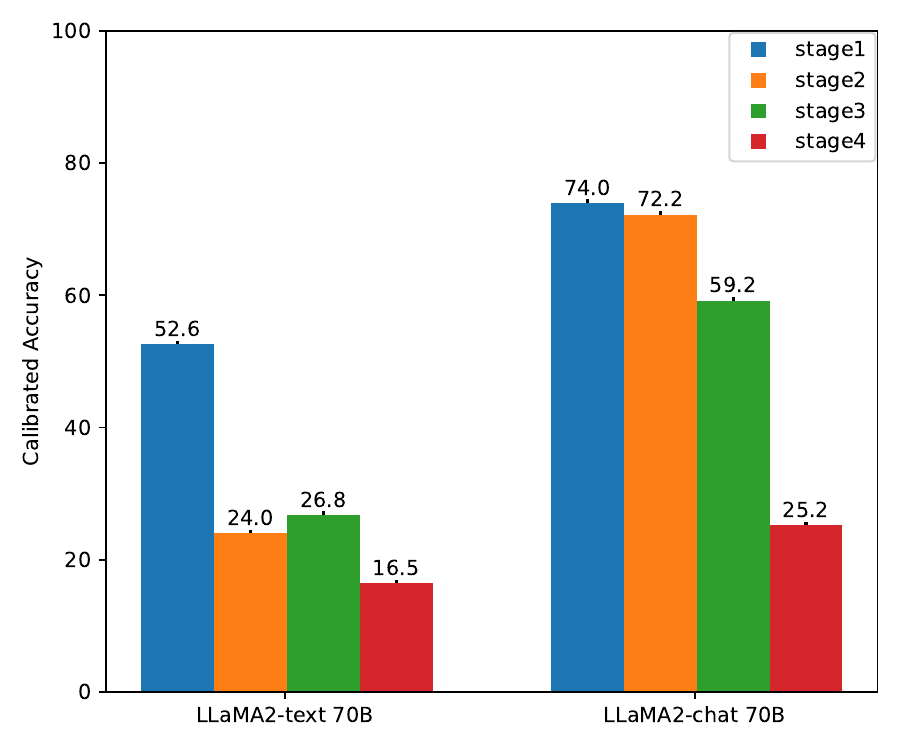}
\end{center}
\caption{Comparison of the performance of Llama-2-70B and Llama-2-chat-70B at each stage on \textsc{CogLM}.}
\label{fig:chat_text}
\end{figure}

\paragraph{Models}
We perform evaluations on the most recent and popular architectures for NLP tasks and restrict our experiments to LLMs. We conduct experiments on the popular family of GPT architecture: OPT series \citep{opt}, including models with sizes of 125M, 1.3B, 2.7B, and 6.7B, optimised for text completion; GPT-3.5-Turbo, optimised for chat; and GPT-4, whose training and architecture details are unknown \citep{gpt4}. We also perform experiments on Llama-2 family of models\citep{llama2}, including models with scale of 7B, 13B and 70B. In particular, Llama-2 series are pretrained generative language models for text completion, while Llama-2-chat is fine-tuned variation optimised for dialogue applications (see Table~\ref{tb:model} for statistics of used LLMs). We conduct experiments on NVIDIA A100 with greedy sampling unless otherwise specified.

\paragraph{Evaluation}
% 这里写prompt以及对于text-completion model和chat-completion model的采样策略
For GPT-3.5-Turbo and GPT-4, we use the Open AI API\footnote{We use "2023-03-15-preview" version for both.} to run all the evaluations.
% As GPT-3.5-Turbo and GPT-4 are chat-completion models, we provide the instructions in chat format. 
For OPT, Llama-2 and Llama-2-chat series models, we use the weights provided on the Huggingface hub\footnote{\url{https://huggingface.co/}}. Llama-2-chat models are used as chat-completion models, while the others are used as text-completion models. For text-completion models, as they lack the ability to follow instructions and their output format is difficult to control, we concatenate each option with the corresponding question as input, and take the option with the highest generation probability as the model's prediction. For chat-completion models, we constrain the format of the model's generated answers through instructions. \footnote{We set the valid output format as: “The answer is \boxed{option}” in the prompt.} We consider a model to provide a valid answer even if the format is incorrect. Unless specified otherwise, we always ask the model to provide a single answer with explanations. The accuracy for a stage is calculated as a macro average of the accuracies of each part of that stage.

% \paragraph{Sampling settings}
% % 温度系数，top_p, 数学表达
% We use greedy sampling across the experiments unless otherwise specified.
% Unless specified otherwise, 
% Running \textsc{CogLM} on other LMs is left for future work.

\subsection{Main Results}
\label{sec:main}
As shown in Table~\ref{tb:main}, We run the model with the largest number of parameters in each series on \textsc{CogLM}, and report the adult human performance for comparison. Overall, the cognitive abilities of OPT, Llama-2-chat-70B, GPT-3.5-Turbo, and GPT-4 models successively increase, and the performance of each model gradually declines with the increase of stage, consistent with humans. Specifically, the latest state-of-the-art model, GPT-4, has demonstrated remarkable cognitive capabilities, achieving a level comparable to that of a 20-year-old human. It is also worth noting that both GPT-3.5-Turbo and GPT-4 surpass humans in empathy ability at stage 2, which is natural, as humans tend to have some degree of selfishness. Despite its superior performance, GPT-4's performance on plan ability (59.4) is still barely satisfactory, far behind that of humans (95.6), which is consistent with the conclusion of \citet{plan}. Our results indicate that enhancing the ability of planning is the major direction for improving the overall cognitive abilities of LLMs in the future. For more detailed evaluation results, please refer to Appendix Table~\ref{tb:main_detailed}.
%and is more suitable as a tool for humans to use.

%\paragraph{What are the similarities and differences between the cognitive development of LLMs and humans?}
\subsection{Analysis and Discussion}
\subsubsection{What are the key factors affecting the cognitive abilities of LLMs?}
 We explore this question from two perspectives: the parameter size and the optimization objective of LLMs, as they are proven to have significant impact on other abilities. We leave the exploration of factors that require changes to the parameters of LLMs (e.g. fine-tuning on different types of datasets) for future work.

\paragraph{The parameter size of LLMs}
As shown in Figure~\ref{fig:parameter}, we compare the overall performance of models with different parameter size across OPT and Llama-2-chat series, and report the performance of humans at different stages as a reference. Specifically, the cognitive abilities of LLMs continuously improve as the size of model parameters increases, which is in line with the conclusion in  \citet{ren2024large}. 
%which follows the same pattern as human cognitive abilities continuously improving with age. 
% 这与...研究中
% We hypothesize that the reason for this phenomenon is that the parameter size of LLMs could be analogous to a child's brain capacity, and cognitive abilities continuously improve with the development of the brain, as stated in \citet{piaget}. 

\begin{table*}[th]
    \centering
    \small
    \renewcommand{\arraystretch}{1.1}
    \renewcommand\tabcolsep{4.5pt}
    \begin{tabular}{l c c c c c c c c c c c}
    \toprule
    Ability&const&early&semio&empat&conse&induc&rever&deduc&propo&plan&Avg\\ \midrule
       Base & \textbf{92.0} & 97.3 & \textbf{90.6} & 85.5 & \textbf{65.9} & 64.0&61.3&27.5&49.0&6.7&64.0\\ 
       $w/$ COT & \textbf{92.0} & 97.3 & 89.3 & 85.5 & 65.7 & 64.0&\textbf{62.3}&26.9&50.5&\textbf{9.8}&64.3\\ 
       $w/$ SC T=0.3 & 91.0 & \textbf{97.6} & 90.3 & \textbf{86.0} & 65.7 & 66.0&61.0&\textbf{27.9}&\textbf{53.5}&4.8&\textbf{64.4}\\ 
       $w/$ SC T=0.7 & 91.0 & \textbf{97.6} & 90.0 & 85.5 & 65.7 & \textbf{66.7}&61.3&27.7&52.0&3.5&63.1\\ 
        \bottomrule
    \end{tabular}
    \caption{Calibrated accuracy of GPT-3.5-Turbo on \textsc{CogLM} with multiple settings. "Base" indicates settings where both COT and SC are not used. }
    \label{tb:ana}
\end{table*}

\paragraph{The optimization objective of LLMs}
As shown in Figure~\ref{fig:chat_text}, we compare the performance of Llama-2-70B and Llama-2-chat-70B at each stage on \textsc{CogLM}. The results show that the performance of both models generally declines with the increase of stage, while the performance of Llama-2-chat-70B far exceeds that of Llama-2-70B at every stage. Given that Llama-2-chat-70B is further fine-tuned on dialogue data and RLHF trained compared to Llama-2-70B, it suggests that LLMs could potentially enhance their cognitive abilities through learning to chat with humans, as RLHF is another approach for LLMs to learn the world, apart from text pretraining.

Based on the two sets of experiments above, we can draw the conclusion that the parameter size and optimization objective are key factors affecting the cognitive abilities of LLMs.

\subsubsection{Can advanced technologies help enhance LLMs' cognitive abilities?}
To answer this question, we applied two representative techniques separately to measure whether cognitive abilities of LLMs could be improved.

\paragraph{Effect of Chain-of-Thought}
\label{sec:effect-cot}
% 通过让语言模型生成思维链，进而解决问题的方式已被证明能够在多数情况下显著提升语言模型的性能。我们好奇COT能否在帮助模型提高认知能力方面取得效果。因此，我们在\textsc{CogLM}上分别测试了 with and without COT下语言模型的性能。

The approach of guiding LLMs to subsequently solve problems has been shown to significantly enhance the performance in most scenarios \citep{COT}. Thus, we are curious whether Chain-of-Thought (COT) can also be effective in improving the cognitive abilities of LLMs. We tested the performance of GPT-3.5-Turbo with and without COT separately on the \textsc{CogLM} and the results are shown in Table~\ref{tb:ana}. From the perspective of the average calibrated accuracy of all the cognitive abilities, COT does not bring a significant performance improvement. We hypothesize that this is because cognitive abilities are inherent to the LLMs and could not be enhanced through multi-step reasoning.

\begin{table*}[t]
    \centering
    \small
    \renewcommand{\arraystretch}{1.0}
    \renewcommand\tabcolsep{4.0pt}
    \begin{tabular}{l c c c c c c c c c c c}
    \toprule
    Erased Ability&const&early&semio&empat&conse&induc&rever&deduc&propo&plan&/\\ \midrule
       GSM8K & 0.1 & 0.6  & 25.5 & 16.6 & 38.6 & 30.7&21.2&12.1&1.0&2.2&59.9\\ 
       StrategyQA & 3.8 & 9.1 & 5.6 & 33.5 & 15.7 & 14.7&18.8&28.4&12.9&31.6&65.2\\ 
        \bottomrule
    \end{tabular}
    \caption{Accuracy (\%) of GPT-3.5-Turbo on GSM8K and StrategyQA datasets when different cognitive abilities are erased. }
    \label{tb:gsm}
\end{table*}

\paragraph{Effect of Self-Consistency}
Self-Consistency (SC) \citep{SC} is another commonly used method that can effectively enhance the performance of LLMs. Multiple candidate predictions to a specific problem are suggested to generate through sampling following with a voting mechanism to eliminate noise introduced by single sampling. We conducted experiments with sampling times as 40 at temperature $T$ of 0.3 and 0.7, respectively. As shown in Table~\ref{tb:ana}, similar to COT, SC can only bring about a very marginal improvement. This phenomenon is consistent with human. For example, for a boy who lacks the ability of empathy, no matter how many times he is asked to choose, he may find it difficult to realize that a scarf might be a more suitable gift for his grandmother than a lollipop. 

Based on the two sets of experiments above, we can draw the conclusion that similar to human beings, it is challenging to achieve significant improvements in LLMs' cognitive abilities without external stimuli.

\subsubsection{How Cognitive Ability Affects the Performance of LLM}
According to PTC, the development of human cognitive abilities is a gradual process, where the cognitive abilities of early stages can influence the advanced cognitive abilities. Additionally, cognitive abilities significantly determines the capacity to solve real-world problems. Therefore, we are very interested in whether these two aspects are similarly manifested in LLMs.

\begin{figure}[t]
\begin{center}
\includegraphics[width=0.44\textwidth]{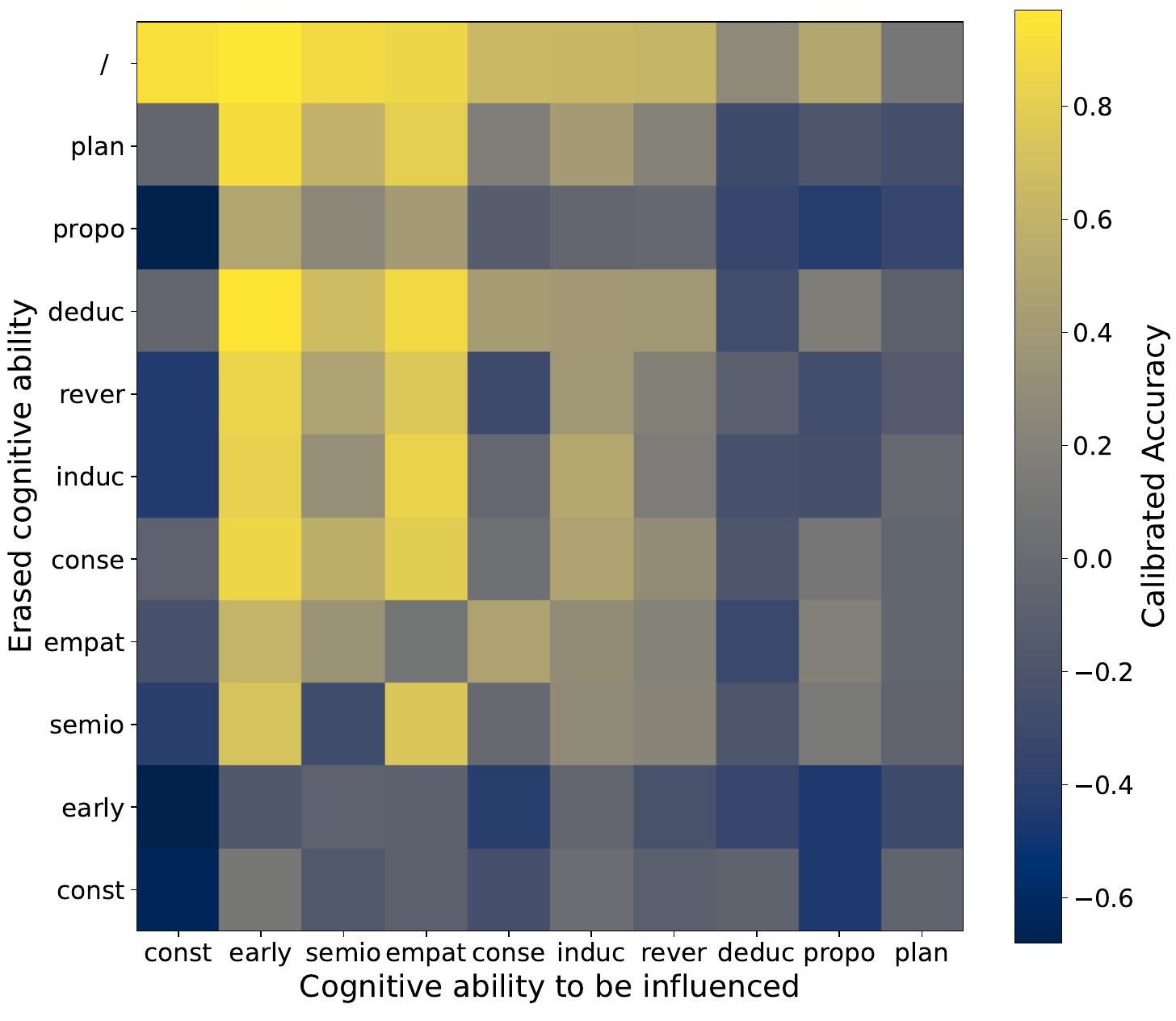}
\end{center}
\caption{Cognitive ability interdependence matrix. The vertical axis represents cognitive abilities erased through prompts, and the color depth (calibrated accuracy) indicates the impact on the corresponding horizontal axis abilities after erasure.}
\label{fig:interdependence}
\end{figure}

\paragraph{The Interdependence Between Cognitive Abilities}
Through preliminary experiments, we found that advanced LLMs' ability to follow instructions can help us erase specific cognitive abilities using a cognitive-ability-setting-prompt (e.g., "You have not yet formed a sense of empathy". See Appendix Table~\ref{table:appendix-prompt} for all the prompts). On this basis, We investigated the interdependence of cognitive abilities in LLMs by selectively removing specific cognitive capabilities and testing them on \textsc{CogLM}. According to the experimental results shown in Figure~\ref{fig:interdependence}, we can draw the following conclusions: (1) Advanced cognitive abilities significantly rely on early cognitive abilities, which indicates that the dependency relationships of LLMs' cognitive abilities are similar to those of humans. (2) The darker colors along the diagonal indicate that the way we erase the corresponding cognitive abilities is effective. (3) Constancy is a fundamental capability (in line with PTC), as it significantly influences and is influenced by other cognitive abilities.

\paragraph{The Dependence of Downstream Ability on Cognitive Ability}
In Table~\ref{tb:main}, we observed a gradual increase in cognitive abilities for OPT, Llama-2, GPT-3.5-Turbo, and GPT-4. On the other hand, based on extensive evaluation studies \citep{bigbench,llama2,liang2022holistic}, we also noted that this ranking result corresponds with the overall performance of LLMs when it comes to solving downstream tasks. This suggests that cognitive abilities are significantly correlated with practical skills for LLMs.
To further understand this correlation, we conducted experiments to assess LLMs' performance on downstream tasks when specific cognitive abilities were erased by cognitive-ability-setting-prompt. We chose representative math reasoning dataset GSM8K \citep{GSM8K} and commonsense reasoning dataset StrategyQA \citep{CSQA} to conduct our experiments. 
As shown in Table~\ref{tb:gsm}, it is reasonable that the erasure of hypothetico-deductive, propositional operation and plan abilities significantly impact the performance of GPT-3.5-Turbo on GSM8K as they are core abilities to solve math problems. We also found that the erasure of other cognitive abilities (especially in early stages) can also bring a strong impact, even if they may not seem helpful in solving math problems. Similar conclusions can be drawn on StrategyQA. These findings indicate that LLMs' abilities to solve downstream tasks is positively correlated with the level of cognitive abilities. 
The advanced cognitive capabilities of GPT-3.5-Turbo and GPT-4 on \textsc{CogLM} partially account for their outstanding performance in various downstream tasks.
From this perspective, we can further understand that Zero-shot COT \citep{zero-cot} is essentially enhancing LLMs' cognitive ability of deduction for improved performance on downstream tasks by incorporating "Let's think step by step" into the prompt.

\begin{table*}[t]
    \centering
    \small
    \renewcommand{\arraystretch}{1.2}
    \renewcommand\tabcolsep{4.5pt}
    \begin{tabular}{l c c c c c c c c c c c}
    \toprule
    Ability&const&early&semio&empat&conse&induc&rever&deduc&propo&plan&Avg\\ \midrule
       $w/o$ COC & 32.0 & \textbf{78.6} & \textbf{74.5} & 54.9 & -0.2 & 41.3&25.3&6.1&-2.0&-0.1&31.0\\ 
       $w/$ COC & \textbf{54.2} & 57.3 & 53.3 & \textbf{69.4} & \textbf{34.1} & \textbf{41.9}&\textbf{34.7}&\textbf{8.1}&\textbf{26.9}&\textbf{1.5}&\textbf{38.1}\\ 
        \bottomrule
    \end{tabular}
    \caption{Calibrated accuracy of Llama-2-chat-7B on \textsc{CogLM} with and without Chain-of-Cognition from GPT-3.5-Turbo as input. }
    \label{tb:coc}
\end{table*}

\subsubsection{Potential applications of advanced LLM cognitive ability}
Although there is still room for improvement, the cognitive abilities of advanced LLMs have approached levels close to that of adult humans as discussed in Section \ref{sec:main}. A natural question is, what are the potential applications for advanced LLMs' cognitive abilities? When humans address cognitive questions, they deduce and provide answers based on their cognitive abilities. While we have demonstrated in Section \ref{sec:effect-cot} that the cognitive chain-of-thought (Chain-of-Cognition, COC) generated by LLMs barely help self-address cognitive questions, we are curious whether COC can assist early-aged LLMs in improving cognitive performance. On this basis, we use the question together with the COCs generated by advanced LLM (GPT-3.5-Turbo) as input to test the performance of early-aged LLM (Llama-2-chat-7B) on \textsc{CogLM}. As shown in Table~\ref{tb:coc}, in most cognitive abilities, COC can significantly improve the performance of Llama-2-chat-7B. We leave the research on using COC from advanced LLMs to guide the improvement of cognitive abilities in early-aged LLMs and even children for future exploration.

%% file: related_work.tex
\section{Related Work}
\paragraph{LLM Evaluation}
Due to the importance of LLMs, their abilities have been thoroughly evaluated on a wide range of problems. Large-scale efforts have been invested in constructing large benchmarks itegrated with numerous LM evaluations across a number of fields \citep{bigbench, liang2022holistic, 2020measuring, biderman2023pythia}. Due to the superior performance of LLMs in a number of traditional NLP tasks, recently challenging tasks have been proposed to test the upper bound performance of LLMs \citep{MATH, plan, abstract_reason}. Some benchmarks include evaluation of specific cognitive abilities, such as common sense reasoning \citep{ismayilzada-etal-2023-crow}, planning \citep{xie2024travelplanner}, and deductive reasoning \citep{saparov2022language}. While previous benchmarks focus on measuring either a type or a category of advanced ability in LLMs, few studies explore the development relationship between different abilities, which is crucial for understanding the emergence of LLMs' abilities.

\paragraph{Cognitive psychology survey on LLMs}
Several works introduce tools from cognitive psychology to study LLMs. Such as understanding the behavior in LLMs \citep{cognitivebias, towardsunstanding, intuition, portelance2023predicting}, exploring the human-like abilities in LLMs \citep{human-like,ToM,TE,mbti}, and improving LLMs' performance on certain task \citep{thinkingoutaloud}. Our work is most similar to present work on using  cognitive psychology to explore whether LMs “learn and think like people” by \citet{usingcognitive}, which suggests that LLMs struggle to reason causally due to the differences in how humans and LLMs learn about the world. The key difference in our approaches is that \citet{usingcognitive} aims to study GPT-3 by assessing its advanced abilities (e.g. decision-making, information search, deliberation, and causal reasoning), while we analyze the relationships between the cognitive abilities of LLMs from the perspective of development, rather than assessing the level of a single cognitive ability of LLMs. Additionally, the other concurrent work \citep{shah2024development} considers the developmental alignment of LLMs' performance during training to the trajectories of children's thinking, primarily measuring the growth trajectories of various cognitive abilities, whereas our measurement of "development" focuses more on the sequence relationships of different cognitive abilities emerging. 
% \citet{mbti} introduce MyersBriggs Type Indicator (MBTI) to explore whether LLMs with human-like abilities exhibit human-like personalities.
% \citep{intuition} test GPT-3.5 using cognitive response tests and find that the LM’s error mode “mirrors intuitive behavior as it would occur in humans in a qualitative sense.”
% explore the similarity between humans and LMs

\paragraph{Piaget's Theory of Cognitive Development}
 Theory of Cognitive Development (PTC) is the most authoritative theory in the development of psychology, developed by Jean Piaget \citep{piaget}. PTC suggests that intelligence grows and develops through a series of stages. As children interact with the world around them, they continually add new knowledge, build upon existing knowledge, and adapt previously held ideas to accommodate new information. PTC is widely used in education, psychology, linguistics, and neuroscience, providing a theoretical framework and methodology for research in these areas. 

 %  Furthermore, PTC focuses not only on understanding how children acquire knowledge, but also on understanding the nature of intelligence \citep{childrenevaluation}. 

%% file: conclusion.tex
\section{Conclusions}
%Large Language Models perform very well on a large range of NLP tasks. Despite the great success, it is important to understand whether LLMs can achieve human-like cognitive abilities or not. 
In this paper, we introduce Piaget's Theory of Cognitive Development (PTC) as a tool to track the development of cognitive abilities of LLMs. We construct \textsc{CogLM} based on the scenerio experiments used in PTC, and conduct thorough human testing to ensure the alignment between \textsc{CogLM} and PTC. Through extensive experiments on multiple series of LLMs, we show that: (1) In our testing framework, Human-like cognitive abilities have emerged in advanced LLMs (such as GPT-4), comparable to those of 20-year-old humans. (2) The parameter size and optimization objective are two key factors affecting the cognitive abilities of LLMs. (3) The ability of performing downstream tasks is positively correlated with the level of cognitive abilities.
We believe that our findings can provide a novel insight into the emergence of abilities in LLMs, and shed light on the future development advanced abilities of LLMs.

%% file: limitation.tex
\section*{Limitations}
Despite obtaining some valuable findings through CogLM, our current exploration does not consider the language model's performance at different training stages to further provide insights for model training, and we will explore it in our future work.

%% file: ethic.tex
\section*{Ethics Statement}
Our dataset does not contain any harmful or offensive contents. Any personal or sensitive information is anonymized and treated with utmost confidentiality. We ensure the protection of participants’ privacy and obtain informed consent for data collection, annotation, and analysis. We incentivized all the annotators uniformly throughout the annotation process.

%% file: acknowledgement.tex
\section*{Acknowledgements}
This work is supported by the Beijing Natural Science Foundation, China (Nos. 4222037, L181010).

%% file: appendix.tex
\section{Appendix}

\begin{table*}[th]
\centering
\small
\begin{tabular}{lccccccccccc}
\toprule
Model & Method & const & early & semio & empat & rever & conse & induc & deduc & propo & plan \\ \midrule
\multirow{2}{*}{OPT-125M} & Concat & -16.0 & 38.7 & 26.3 & -19.2 & 4.0 & 16.2 & 9.3 & 5.6 & 4.0 & 2.2 \\
 & Option & -18.1 & 0.0 & 10.3 & -36.9 & 4.0 & -18.0 & -4.0 & 2.9 & -3.5 & -6.0 \\ \midrule
\multirow{2}{*}{GPT-2} & Concat & -20.0 & 41.3 & 30.4 & -4.7 & 5.3 & 10.8 & 9.3 & 4.5 & -0.5 & 4.8 \\
 & Option & -25.0 & -6.7 & 16.7 & -38.5 & 5.3 & -18.1 & -6.7 & 0.8 & -0.5 & -6.0 \\ \bottomrule
\end{tabular}
\caption{Performance Comparison of "Concat" and "Option" Testing Methods on \textsc{CogLM} Using GPT-2 and OPT-125M.}
\label{tb:test method}
\end{table*}

\subsection{Testing Method of Text-completion Models}

For text-completion models, as they lack the ability to follow instructions and their output format is difficult to control, we concatenate each option with the corresponding question as input, and take the option with the highest generation probability as model prediction. When calculating the generation probability, we normalized the generation length to eliminate the influence of the option length. Additionally, we compared the approach of having the model interpret the questions as multiple choice and using a letter as the concatenated answer (denoted as "option") with our existing testing method (denoted as "concat") using GPT-2 and OPT-125M. The results (Table \ref{tb:test method}) show that changing the testing method from "concat" to "option" leads to a significant decrease in the performance of the text-completion model. We suppose this is due to the text-completion model being more sensitive to factors such as position bias and model preference compared to the chat-completion model. As a result, directly concatenating the options with the question and ranking them based on probability is more suitable for testing the text-completion model.

% Please add the following required packages to your document preamble:
% \usepackage{multirow}

\begin{table*}[th]
\renewcommand{\arraystretch}{1.2}
\centering
\small
\setlength{\tabcolsep}{1.99pt} % 设置列之间的空白长度
\begin{tabular}{lccp{0.001pt}ccp{0.001pt}cccp{0.001pt}ccccc}
\toprule
\multicolumn{1}{c}{\multirow{2}{*}{Model}} & \multicolumn{2}{c}{stage1} && \multicolumn{2}{c}{stage2} && \multicolumn{3}{c}{stage3} && \multicolumn{3}{c}{stage4} & \multirow{2}{*}{Acc} & \multirow{2}{*}{Age} \\ \cline{2-3}\cline{5-6}\cline{8-10}\cline{12-14}
\multicolumn{1}{c}{} & const & early && semio & empat && conse & induc & rever && deduc & propo & plan &  \\ \hline
OPT 125M & -16.0 & 38.6 && 26.3 & -19.0 && 16.2 & 9.3 & 4.0 && 5.6 & 4.0 & 2.2 &  7.1 & 4.8 \\ 
OPT 1.3B & -20.0 & 52.0 && 38.4 & -11.0 && 1.0 & 12.0 & 13.3 && -6.1 & 2.5 & -3.0 & 7.9 & 5.2 \\ 
OPT 2.7B & -8.0 & 53.3 && 43.7 & -9.5 && 9.4 & 12.0 & 21.1 && -1.3 & 5.5 & 3.4 &  12.95 & 6.1 \\ 
OPT 6.7B & -4.0 & 64.2 && 41.1 & -3.0 && 13.5 & 10.6 & 20.1 && -0.8 & -0.5 & 14.2 &  15.5 & 6.5 \\ 
LLaMA2-text 7B & 16.0 & 82.6 && 43.7 & -4.0 && 20.0 & 1.3 & 24.0 && -16.8 & 14.5 & 24.4 & 20.5 & 7.3 \\ 
LLaMA2-text 13B & 28.0 & 84.0 && 42.4 & -3.0 && 13.0 & 13.5 & 24.0 && -15.7 & 35.5 & 13.0 & 23.4 & 7.7 \\ 
LLaMA2-text 70B & 52.1 & 96.2 && 78.5 & 66.2 && 68.4 & 65.3 & 44.0 && 15.2 & 40.0 & 20.6 & 54.6 & 14.1 \\ 
LLaMA2-chat 7B & 32.0 & 78.6 && 74.5 & 54.9 && -0.2 & 41.3 & 25.3 && 6.1 & -2.0 & -0.1 & 31.04 & 10.3 \\ 
LLaMA2-chat 13B & 44.0 & 89.3 && 78.5 & 56.5 && 16.2 & 42.6 & 32.0 && -0.2 & 17.5 & 1.5 & 37.8 & 11.35 \\ 
LLaMA2-chat 70B & 52.1 & 96.2 && 78.5 & 66.2 && 68.4 & 65.3 & 44.0 && 15.2 & 40.0 & 20.6 & 54.6 & 14.1 \\ 
GPT-3.5-Turbo & 92.0 & \underline{97.3} && 90.6 & \underline{85.5} && 65.9 & 64.0 & 61.3 && 27.5 & 49.0 & 6.7 & 64.0 & 16.1 \\ 
GPT-4 & \underline{96.0} & \underline{97.3} && \underline{96.0} & \textbf{90.3} && \underline{90.4} & \underline{78.7} & \underline{78.7} && \underline{99.4} & \underline{61.0} & \underline{59.4} & \underline{84.7} & \underline{20.0} \\ 
Human & \textbf{100.0} & \textbf{98.0} && \textbf{96.1} & 84.2 && \textbf{98.2} & \textbf{91.6} & \textbf{92.0} && \textbf{100.0} & \textbf{82.0} & \textbf{95.6} & \textbf{93.7} & \textbf{21.5} \\
\bottomrule
\end{tabular}
\caption{Calibrated accuracy (\%) of all models in evaluating series. Acc and Age refer to calibrated accuracy and the age of equivalent human performance. The value of Age is calculated according to Equation~\ref{equation: L_inter} and rounded to the nearest integer. Bold indicates the best performance.}
\label{tb:main_detailed}
\end{table*}

\begin{table*}[th]
\small
\centering
\begin{tabular}{p{13.5cm}}
\toprule
\textbf{\textit{Constancy}} \\
Please imagine yourself as a child aged 0-2 years old. According to Piaget's theory of cognitive development, you are currently unable to recognize that objects exist both within and outside the field of vision and maintain a certain level of stability. \\
\midrule
\textbf{\textit{Early Representation}}\\
Please imagine yourself as a child aged 0-2 years old. According to Piaget's theory of cognitive development, You currently cannot give objects corresponding meanings, nor do you have a definite perception of permanent objects in the universe. \\
\midrule
\textbf{\textit{Semiotic Function}} \\
Please imagine yourself as a child aged 2-7 years old. According to Piaget's theory of cognitive development, You are currently unable to use symbols to represent things and concepts. \\
\midrule
\textbf{ \textit{Empathy} } \\
Please imagine yourself as a child aged 2-7 years old. According to Piaget's theory of cognitive development, You are accustomed to thinking from your own perspective and have not yet formed a sense of empathy. \\
\midrule
\textbf{ \textit{Reversibility} } \\
Please imagine yourself as a child aged 7-11 years old. According to Piaget's theory of cognitive development, You are currently unable to understand the reversibility of physical operations and unable to reverse thinking. \\
\midrule
\textbf{ \textit{Conservation}} \\
Please imagine yourself as a child aged 7-11 years old. According to Piaget's theory of cognitive development, You think that external changes in form (length, shape, etc.) may affect the basic properties of an object (mass, volume, etc.). \\
\midrule
\textbf{ \textit{Induction}} \\
Please imagine yourself as a child aged 7-11 years old. According to Piaget's theory of cognitive development, You currently cannot infer universal rules based on observed results. \\
\midrule
\textbf{ \textit{Hypothetico-Deductive}} \\
Please imagine yourself as a teenager aged 11-18 years old. According to Piaget's theory of cognitive development, You are currently unable to deduce practical problems based on specific assumptions or rules. \\
\midrule
\textbf{ \textit{Propositional Operation}} \\
Please imagine yourself as a teenager aged 11-18 years old. According to Piaget's theory of cognitive development, You are currently unable to understand propositions and determine the logical relationships between propositions. \\
\midrule
\textbf{ \textit{Plan} } \\
Please imagine yourself as a teenager aged 11-18 years old. According to Piaget's theory of cognitive development, You are are currently unable to develop solutions based on specific problem. \\
\bottomrule
\end{tabular}
\caption{Cognitive-ability-setting-prompts of different cognitive abilities.}
\label{table:appendix-prompt}
\end{table*}

\begin{table*}[th]
% \small
\renewcommand{\arraystretch}{0.51}
\centering
\begin{tabular}{p{13.5cm}}
\toprule
\textbf{Question:} How many main stages are included in Jean Piaget's cognitive development theory? \\  \textbf{Answer:} 4 \\
\midrule
\textbf{Question:} Which stage in Piaget's theory marks the point at which children are capable of logical thinking and understanding concepts like quantity, category, space, and time? \\  \textbf{Answer:} Formal operational stage \\
\midrule
\textbf{Question:} What type of operations can children perform during the concrete operational stage? \\  \textbf{Answer:} Addition and subtraction \\
\midrule
\textbf{Question:} Janie knows that a bird has wings and can fly. While camping she finds a bat and thinks it's a bird, but realizes that it doesn't act the same way as a bird. She is confused. She is using what adaptation process with this new knowledge? \\  \textbf{Answer:} Accommodation \\
\midrule
\textbf{Question:} What kind of activities can children engage in during the formal operational stage? \\  \textbf{Answer:} Abstract thinking and logical reasoning \\
\midrule
\textbf{Question:} In Jean Piaget's cognitive development theory, which stage marks the point at which children begin to use symbols and language to represent objects? \\  \textbf{Answer:} Preoperational stage \\
\midrule
\textbf{Question:} Which of the following is NOT one of Piaget's stages of cognitive development? \\  \textbf{Answer:} Abstract operational stage \\
\midrule
\textbf{Question:} Children in the concrete operational stage typically understand what type of concepts? \\  \textbf{Answer:} Concepts of quantity and space \\
\midrule
\textbf{Question:} What types of problems can children in the formal operational stage handle? \\  \textbf{Answer:} Abstract and hypothetical problems \\
\midrule
\textbf{Question:} What are common characteristics of children in the preoperational stage? \\  \textbf{Answer:} Subject to egocentrism \\
\midrule
\textbf{Question:} In the sensorimotor stage, how do children primarily explore the world? \\  \textbf{Answer:} Sensation and movement \\
\midrule
\textbf{Question:} In the sensorimotor stage, how do infants primarily interact with the world? \\  \textbf{Answer:} Observation and sensation \\
\midrule
\textbf{Question:} Jean Piaget's cognitive development theory primarily focuses on which age group of children? \\  \textbf{Answer:} Infants and children \\
\midrule
\textbf{Question:} What types of problems can children in the concrete operational stage typically understand? \\  \textbf{Answer:} Logical problems \\
\midrule
\textbf{Question:} What characteristics do children in the formal operational stage exhibit? \\  \textbf{Answer:} Ability to engage in abstract thinking and hypothetical reasoning \\
\midrule
\textbf{Question:} What does Jean Piaget's cognitive development theory emphasize? \\  \textbf{Answer:} The active role of individuals in cognitive development \\
\midrule
\textbf{Question:} In Jean Piaget's cognitive development theory, which stage marks the point at which children can engage in abstract thinking and hypothetical reasoning? \\  \textbf{Answer:} Formal operational stage \\
\midrule
\textbf{Question:} What does Piaget's theory emphasize as influencing cognitive development? \\  \textbf{Answer:} A balance of social factors and genetic factors \\
\midrule
\textbf{Question:} What can children in the formal operational stage consider when thinking? \\  \textbf{Answer:} Future and hypothetical situations \\
\midrule
\textbf{Question:} What is the primary focus of the sensorimotor stage in Piaget's theory? \\  \textbf{Answer:} Sensory and motor exploration \\
\bottomrule
\end{tabular}
\caption{Examination paper to ensure the annotators are qualified.}
\label{table:appendix-exam}
\end{table*}